# Training Algorithm for Neuro-Fuzzy Network Based on Singular Spectrum Analysis

Yulia S. Maslennikova [a], Vladimir V. Bochkarev [a]

*[a]Kazan federal university, Kremlyovskaya, 18, Kazan, 420008, Russia*
*E-mail address:* yulia.maslennikova@tgtoil.com, vbochkarev@mail.ru

**Abstract**

In this article, we propose a combination of an noise-reduction algorithm based on Singular Spectrum Analysis (SSA) and a standard feedforward neural prediction model. Basically, the proposed algorithm consists of two different steps: data preprocessing based on the SSA filtering method and step-by-step training procedure in which we use a simple feedforward multilayer neural network with backpropagation learning. The proposed noise-reduction procedure successfully removes most of the noise. That increases long-term predictability of the processed dataset comparison with the raw dataset. The method was applied to predict the International sunspot number RZ time series. The results show that our combined technique has better performances than those offered by the same network directly applied to raw dataset.

## 1. Introduction

For many years, the field of time series forecasting was largely analyzed by various statistical approach (see e.g. [1]). Over the past 20 years in the field of nonlinear dynamical system, an artificial neural network (ANN) theory has gained more importance. In comparison with commonly used linear regression algorithms such as ARX, ARIMA [2] this approach has several advantages. According to the Universal approximation theorem the standard multilayer feed-forward network with a single hidden layer, which contains finite number of hidden neurons, is a universal approximator among continuous functions on compact subsets of $R_n$, under mild assumptions on the activation function. [3]. However, the problem of overfitting might occur in cases when network learning was performed for too long or training datasets are not representative enough. In this case neural network model describes random error of noise instead of the underlying relationship [4]. To solve the problem of overfitting, various approaches based on training data preprocessing was proposed. In many articles, the neural network approach combined with Principal Component Analysis (PCA) or Singular Spectrum Analysis (SSA) is used for the time series prediction. ANN long-term prediction model based on SSA was reported by Gholipour *et al.* [5]. The proposed method consists of locally linear neuro fussy (LLNF) model that is optimized for each of the principal components obtained from SSA. The multi-step predicted principal components are combined to reconstruct the prediction of time series. Similar method based on SSA and fussy descriptor models has been applied to predict two geomagnetic activity indices: geomagnetic *aa* and solar wind speed of the solar wind index. [6]. The results demonstrate that the proposed method predicts solar activity time series more accurately then other methods. Notwithstanding the wide application of this approach, there are some inherent drawbacks. One drawback of the combined model is that it consists of a large number of optimizing coefficients.

In this paper, we address the well-known task of short and noisy time series prediction. Basically, the algorithm which we propose in this paper consists of two different steps: data preprocessing based on the

SSA filtering method and step-by-step training procedure in which we use a simple feedforward multilayer neural network with backpropagation learning. The proposed noise-reduction procedure successfully removes most of the noise. That increases long-term predictability of the processed dataset comparison with the raw dataset. We compare the performances of our approach to those of an analogous neural network predictor trained with raw data. The method was applied to predict the International sunspot number RZ time series. In section 2 we shortly explain the theoretical background of SSA. Next in section 3, we consider the basic assumption of our approach for step-by-step training procedure. The proposed model is used to predict the sunspot number time series and to show the performance of this method in comparison with other methods. Finally, a brief conclusion is given in section 4.

## 2. Singular Spectrum Analysis

Singular Spectrum Analysis (SSA) is a general approach to time series analysis and forecast. Algorithm of SSA is similar to that of Principal Components Analysis (PCA) of multivariate data. In contrast to PCA which is applied to a matrix, SSA is applied to a time series and provides a representation of the given time series in terms of eigenvalues and eigenvectors of covariance matrix. Vautard, Yiou, and Ghil [7] describe SSA method. For a standardized time series $x_i$, where sample index $i$ varies from 1 to $N$, and a maximum lag (or window size) $M$, a Toeplitz lagged correlation matrix (each diagonal has a uniform value), is formed by

$$c_j = (N-j)^{-1} \sum_{i=1}^{N-j} x_i\, x_{i+j},\ 0 \le j \le M-1 \qquad (1)$$

The eigenvalues, $\lambda_k$, and eigenvectors (or empirical-orthogonal functions), $E_i^k$, of this matrix are determined and sorted in descending order of $\lambda_k$, where indices $j$ and $k$ vary from 1 to $M$. The $k^{th}$ principal component is

$$a_i^k = \sum_{j=1}^{M} x_{i+j} E_j^k,\ 0 \le j \le N-M \qquad (2)$$

Each component of the original time series indented by SSA can be reconstructed, with the $k^{th}$ reconstructed component (RC) series given by

$$x_i^k = M^{-1} \sum_{j=1}^{M} a_{i-j}^k E_j^k,\ M \le i \le N-M+1 \qquad (3)$$

Expressions for $x_i^k$ for $i < M$ and $i > N - M + 1$ are given by Vautard, Yiou, and Ghil [7]. The fraction of the total variance of the original time series (equal to one for standardized time series) contained in the $k^{th}$ RC is $\lambda_k$, so that, with the sorting used, the RCs are ordered by decreasing information about the original time series. Most of the variance is contained in the first several RCs and most or all of the remaining RCs contain noise. SSA typically decomposes a time series into RCs that are nearly periodic with periods less than M and one or two RCs contain variations in the time series with periods greater than $M$. A pair of RCs with similar $\lambda_k$ typically represents each period less than *M* with significant energy in the original time series [7].

As the final step, the time series $\tilde{x}^{(p)}$ is reconstructed by combining the associated *p* principal components $x_i^k$ :

$$\tilde{x}_i^{(p)} = \sum_{k=1}^{p} x_i^k \qquad (4)$$

The singular spectrum plot (the logarithmic scale plot of singular values of the covariance matrix in decreasing order) can be used to analyze of principal components. To obtain adaptive noise filtering and enhance the Signal-Noise Ratio (SNR) the principal components related to the lower singular values can be omitted in the reconstruction stage (Eq. 4). On the one hand, if all the components are used in reconstructing time series, no information is lost. On the other hand, using the components that have characteristics like noise, which cause the components unpredictability over the long term, reduce the performance of the long-term predictor. Therefore, it is better to consider these components as noise, and assume that these components provide little information on the original time series [7].

## 3. Proposed Method and Results

The multilayer feedforward network can be trained for function approximation (nonlinear regression) or pattern recognition. The process of training a neural network involves tuning the values of the weights and biases of the network to optimize network performance, as defined by the network performance function. The most commonly used performance function for feedforward networks is mean square error (MSE) - the average squared error between the network outputs $a_i$ and the target outputs $t_i$ [4].

For training multilayer feedforward networks, any standard numerical optimization algorithm can be used to optimize the performance function, but there are a few key ones that have shown excellent performance for neural network training. The gradient and the Jacobian are calculated using the technique called the backpropagation algorithm, which involves performing computations backward through the network [4]. The traditional backpropagation algorithm modifies the weights of a neural network in order to find a global minimum of the error function. Therefore, the gradient of the error function is calculated with respect to the weights in order to find the root. Particularly, the weights are modified by going in the opposite direction of the partial derivatives until a global minimum is reached. In the case when the time series have complicated dynamics the error function has many local minima. Therefore the error function is perceived as an intricate problem. To improve the error function minimization procedure we proposed the following algorithm. As a first step, the neural network is trained using simplified version of the time series. This time series is reconstructed using first two principal components. Then training dataset is made accurate using the principal components related to the lower singular values. Thus, we get an opportunity to take into account different features of the time series dynamics. During optimizing this prediction model we use only one simple feedforward multilayer neural network with backpropagation learning. Fig.1 shows the block diagram of proposed algorithm. Optimizing the neurofuzzy model for sequentially reconstructed time series is obtain using the proposed learning algorithm on separate training and validation set. The validation set is randomly chosen as a 10% part of the total data set. This step-by-step learning algorithm performs well in prediction applications. One of the most parameter should be defined before running the algorithm: the embedding dimension *m* (number of regressors from time series as the input to the neural network model). The well-known Takens' theorem gives a sufficient condition as a lower bound for the embedding dimension $m$ f chaotic systems [8] and what output is supposed to give for specific inputs.

The algorithm of long-term prediction is as follows:
1. Construction of the time series trajectory matrix. Decomposition of the time series to principal components via SSA
2. Reconstruction of "simplified version" of the time series using first two principal components.
3. Learning of the feedforward network using "simplified version" of the time
4. Step-by-step reconstruction of the time series using the principal components related to the lower singular values.
5. Finally, learning the feedforward network using raw dataset.

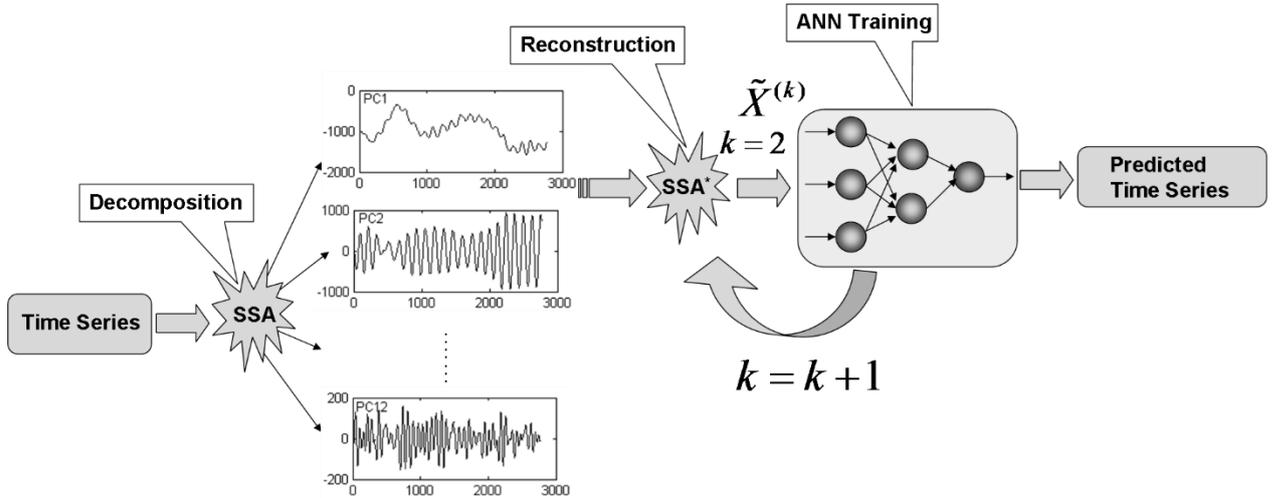

Fig. 1. The proposed method consists of five steps: (1) decomposition of the time series to nonlinear and periodic principal components using SSA, (2) reconstructing the time series using first two principal components, (3) training of the neural network using the reconstructed time series, (4) reconstructing the time series using another principal components and return to step 3.

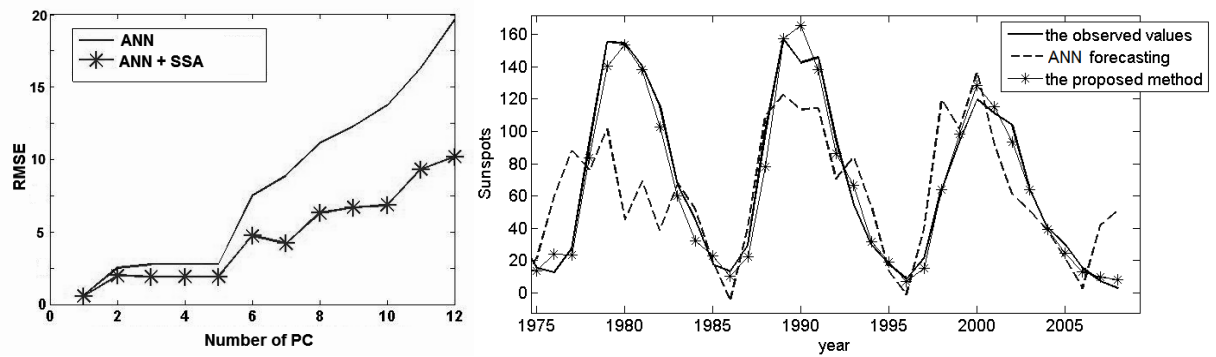

Fig. 2. (a) Number of PC dependence for the training error; solid line with asterisks: combined learning algorithm ANN+SSA; solid line: an analogous neural network predictor trained with raw data;
(b) Long-term prediction of solar cycles 21, 22, 23 (from training dataset); solid line: the observed values, solid line with asterisks: 6-year predicted values, dashed: values predicted by the same network directly applied to raw data.

The proposed combined method was used for long-term prediction of sunspot number time series. For this time series the window size $M$ of lag matrix has been chosen as 35 to cover at least three solar cycles. Embedding dimension of sunspot number time series has been estimated and approximately equal to $m=5$. As a first step, the neural network was trained using the time series, which was reconstructed using first two principal components. The dependence of learning error against the number of principal components whish was used for reconstruction and training is proposed on the Fig. 2a. Fig. 2b shows the long-term prediction of solar cycles 21, 22, 23. It is shown that proposed prediction model is capable of issuing very accurate long-term prediction of sunspot number time series. The peak of the next solar maximum is predicted to be near 88 in 2013-2014 (see Fig. 3).

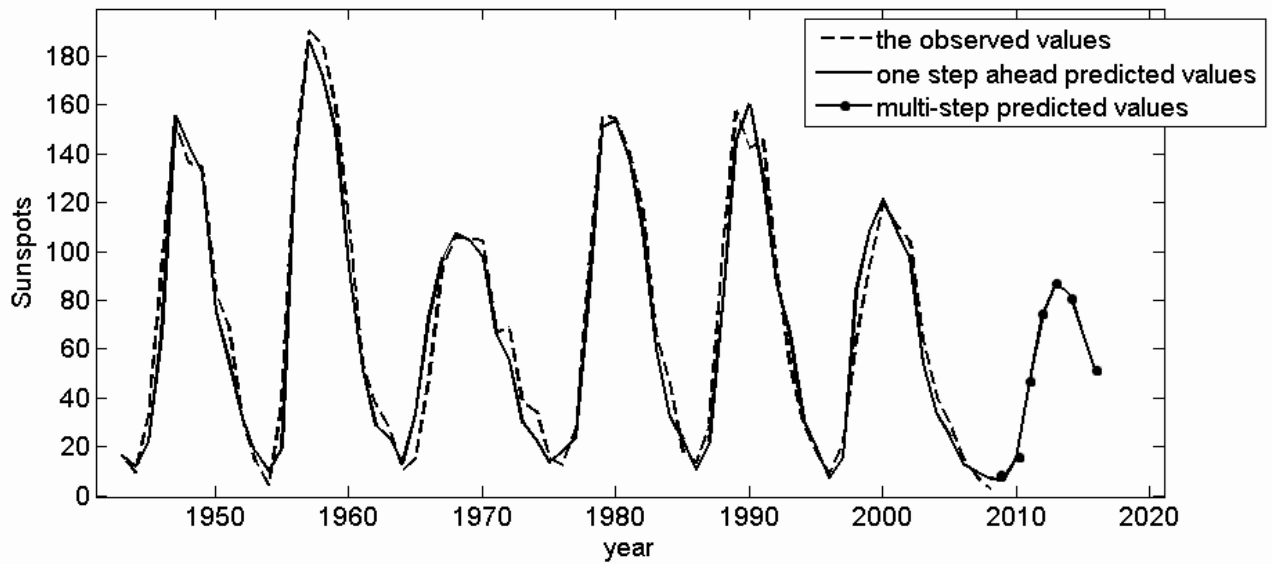

Fig. 3. Long-term prediction of solar cycle 24 by proposed method; dashed: the observed values, solid line from 1944 to 2009: one step ahead predicted values, and solid line with dots from 2009 to 2015: multi-step predicted values.

## 4. Conclusion

The long-term prediction of natural phenomena with a limited number of observations is usually difficult. Here, we have used a combined model for based on feedforward neural network and singular spectrum analysis for the long-term prediction of sunspot time series. The long-term predictions by proposed method are superior in comparison to the same neural prediction model directly applied to raw data for the last three solar cycles, and the peak of the next solar maximum is predicted to be near 88 in 2013-2014.